# ConvSRC: SmartPhone based Periocular Recognition using Deep Convolutional Neural Network and Sparsity Augmented Collaborative Representation

Amani Alahmadi[a,], Muhammad Hussain[a], Hatim Aboalsamh[a] and Mansour Zuair[a]

[a]King Saud University, College of Computer and Information Sciences, Riyadh 11543, Saudi Arabia

## ABSTRACT

Smartphone based periocular recognition has gained significant attention from biometric research community because of the limitations of biometric modalities like face, iris etc. Most of the existing methods for periocular recognition employ hand-crafted features. Recently, learning based image representation techniques like deep Convolutional Neural Network (CNN) have shown outstanding performance in many visual recognition tasks. CNN needs a huge volume of data for its learning, but for periocular recognition only limited amount of data is available. The solution is to use CNN pre-trained on the dataset from the related domain, in this case the challenge is to extract efficiently the discriminative features. Using a pertained CNN model (VGG-Net), we propose a simple, efficient and compact image representation technique that takes into account the wealth of information and sparsity existing in the activations of the convolutional layers and employs principle component analysis. For recognition, we use an efficient and robust Sparse Augmented Collaborative Representation based Classification (SA-CRC) technique. For thorough evaluation of ConvSRC (the proposed system), experiments were carried out on the VISOB challenging database which was presented for periocular recognition competition in ICIP2016. The obtained results show the superiority of ConvSRC over the state-of-the-art methods; it obtains a GMR of more than 99% at FMR = $10^{-3}$ and outperforms the first winner of ICIP2016 challenge by 10%.

## 1. Introduction

Biometric recognition has emerged as a topic of research interest in the past decade. With the development of smart mobile devices equipped with digital cameras and significant computational power, the research has turned its focus to mobile based biometrics.

Face recognition is one of the most popular biometrics, which performs well in controlled environment. However, the performance of face recognition systems declines if the face is partially hidden [1]. The whole face of criminals often does not appear in surveillance videos. In different situations, face is either covered by helmets, hair, glasses or skiing masks. Furthermore, due to cultural and religious reasons in some countries, women cover their faces partially. In most of these cases, the region around the eyes (periocular) is the only visible trait which can be used as biometric (see Figure 1).

Moreover, the acquisition of the image of periocular region does not require high user cooperation and close capture distance unlike other ocular biometrics (e.g., iris, retina, and sclera). The periocular region gives a trade-off between the whole face and the iris alone. Containing the eye and its immediate vicinity, it covers eyelids and eyelashes, nearby skin area and eyebrows (see Figure 2).

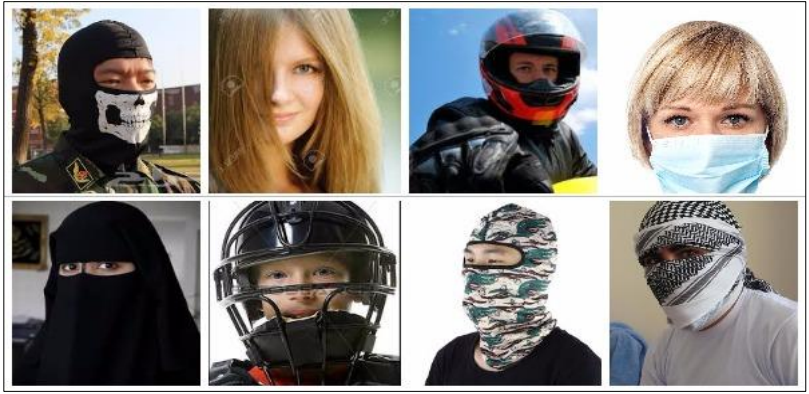

**Figure 1.** Examples of some cases where using periocular biometric is efficient.

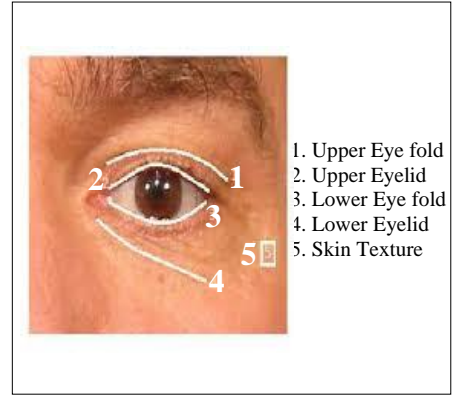

**Figure 2.** An Example of a Periocular Region of an Eye.

Several efforts have been made on periocular recognition, which have been reviewed and discussed in [23, 24, 25]. Most of the existing techniques for periocular recognition use hand-crafted features. Among these techniques are global descriptors such as Block HOG [2], HOG [3] and Color histogram [4], and local descriptors such as SURF [5], LBP [6] and SIFT [7]. The state-of-the-art techniques include the deep sparse representation [8], BSIF [9], Local phase Quantization (LPQ) features [10], Block BSIF [11] and Deep Sparse Filters [22]. Only few works used leaned features [32, 21]. For recognition, mostly KNN with distance metrics such as city block [6, 26, 27], Euclidean [1,28], chi-square [29,30,7] and mean square error [31] has been used. Some works used learning based techniques like SVM [32, 33], and Neural Networks [21, 22].

Recent studies have shown that deep learned features are more effective than hand-crafted features for periocular recognition tasks [23, 16]. In the recent years, though learned features using deep Convolutional Neural Networks (CNN) have attracted a lot of attention in visual recognition tasks due to their outstanding performance [12], a huge volume of data is needed for their learning. However, for periocular classification a huge amount of data is not available and it is hard to learn descriptive features using deep CNNs with a dataset of small size. It has been found in recent researches that a deep CNN, pre-trained on a large dataset from a related domain, can be employed to build an image representation for a visual classification problem with small dataset and delivers impressive performance [13]. In this case, the key question is how to extract features using a pre-trained CNN model for periocular recognition. Most of the methods based on pre-trained deep CNN models used the activations of a fully connected layer as a representation (*global features*). There is a wealth of information encoded in convolutional layers, however, only few works used the activations of the convolutional layers [17, 18]. A convolutional layer encodes the *local features* and preserves the spatial information, in contrast with the holistic features extracted from a fully connected layer.

Recently, Sparse representation based classification (SRC) has shown promising results for different applications [34]. It represents a test sample as a sparse linear combination of the training samples of all classes and then classifies this sample to the class with minimum representation error. Although, it is a common belief that the key success of such classification method is due to the sparseness of the representation, recently, it has been found that it is the collaborative representation, not the sparseness, which is the reason for the effectiveness of this scheme. In a recent work [15], it has been shown that the sparse representation cannot be completely ignored, and the augmentation of a dense collaborative representation with a sparse representation improves the recognition performance. Based on this observation, Naveed et al. [15] introduced Sparsity Augmented Collaborative Representation based Classification (SA-CRC) technique for recognition. In this study, we employed this technique for particular recognition and examined tits usefulness for this problem.

Motivated by the effectiveness and outstanding performance of CNN in many visual recognition tasks, we propose a technique for extracting local features using pre-trained deep CNN model, and a periocular recognition method based on these deep learned features and SA-CRC, we call this method as ConvSRC. The development of this method raises some questions: (1) how to extract discriminative deep features from convolutional layers of a pre-trained CNN model for periocular recognition? (2) can deep features extracted using deep CNN models pre-trained on a dataset from a related domain (e.g. face dataset) be generalized to periocular recognition task? (3) are the features extracted using deep CNN model pre-trained on a dataset of natural images (e.g. ImageNet dataset) effective for this task? (4) which features (i.e. local or global) are more suitable for this task? (5) Is sparse collaborative representation based classification is effective with deep learned features?

To answer these questions, we thoroughly analysed the performance of different feature representations extracted from different convolutional and fully connected layers of the pre-trained VGG-Face [14], which is learned on a dataset from a closely related domain, and VGG-Net trained on ImageNet dataset. For classification, we investigated the impact of using SA-CRC [15] compared to the baseline KNN. Extensive experiments conducted on VISOB, the ICIP2016 challenge dataset for smartphone periocular recognition [16], show that ConvSRC gives promising results; it outperforms the state of the art methods and the recent method reported as the first winner in the ICIP2016 challenge [16]; it outperforms (up to 10%) the first winner.

The main contributions of the paper are summarized below:
- Proposed a simple and effective technique for extracting local features from a pertained deep CNN model, which is effective in periocular recognition.
- Proposed an efficient periocular recognition method (ConvSRC) which is based on local deep CNN features and SA-CRC, and is unsupervised in the sense that deep CNN model has not been trained using periocular dataset.
- Demonstrated that for periocular recognition problem, local and global deep CNN features, even with simple KNN classifier, result in better performance than the state-of the-art hand-crafted and learned features.
- Thoroughly examined the effectiveness of local and global features and shown that for periocular recognition, local features extracted from the last convolutional layer encode more discriminative representation than global features from a fully connected layer.
- Shown that deep CNN features extracted using VGG-Face model pre-trained on face dataset are effective in periocular recognition.
- Found that VGG-16 pre-trained on ImageNet [35] gives results that are almost similar to those obtained using VGG-Face model. This reveals that VGG-Net model learned on natural image dataset is equally effective for periocular recognition.
- Demonstrated that comparing with KNN, SA-CRC significantly improves the recognition results.
- Shown that ConvSRC performs well for both verification and identification.

The rest of the paper is organized as follows. Section 2 provides the detail of the proposed method. The evaluation protocol is described in Section 3. The model selection has been discussed in Section 4. The details of the experiments conducted to show the effectiveness of local and global features are given in Section 3, while the results and discussions are presented in Section 4. Finally, Section 5 concludes the paper.

## 2. The Proposed Methods - ConvSRC

A systematic diagram of the method is shown in Figure 3. First, the input image is resized to 224x224 and passed to VGG Net, which extracts deep CNN features (global /local). Next, the dimension of the feature vector is reduced using PCA. Finally, SA-CRC is used to give the decision. The detail is given in the following subsections.

### 2.1. Deep CNN Features

For extracting deep CNN features, the first task is to select the suitable CNN model. Out of different CNN structures, VGG Net is one of the commonly used model [36], which have shown promising results for different Computer Vision applications. Initially, we adopted very deep VGG-Face model, which is a VGG Net model and is trained on face image dataset that is a related domain. Our selection is based on the fact that periocular region is a part of face image and CNN learns hierarchy of features in such a way that higher level layers encode object parts.

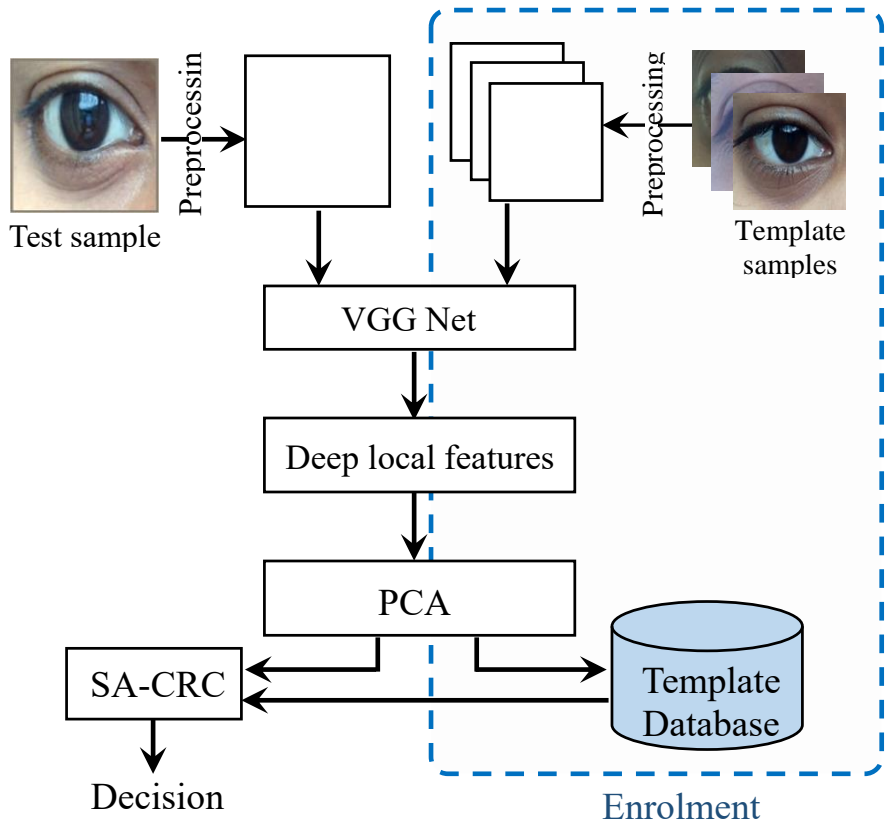

**Figure 3.** The framework of ConvCRC.

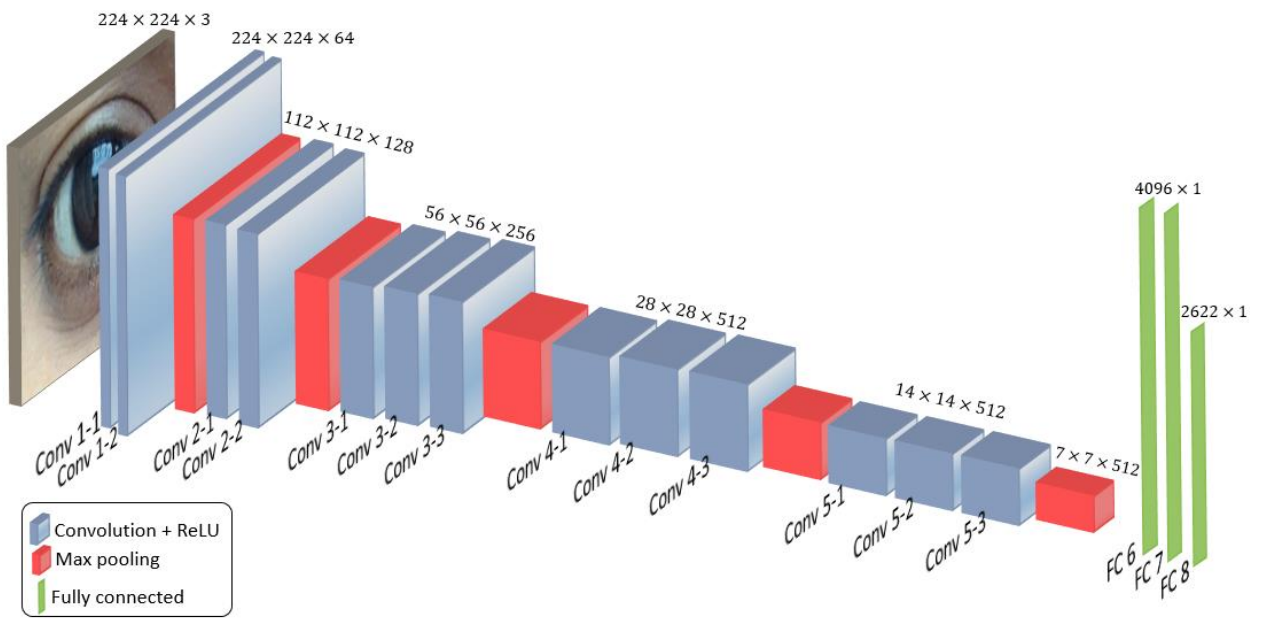

**Figure 4.** The architecture of VGG-Faces.

The structure of very deep VGG-Faces [14] model is shown in Figure 4; it comprises 5 blocks of convolutional (CONV) layers, each followed by max pooling layer and three fully connected (FC) layers. Each CONV layer generates feature maps (activations) by convolving the input with a bank of linear filters, learned during training and is followed by a rectification layer (ReLU), which applies ReLU non-linearity on the features maps. The feature maps of low CONV layers are large in size but small in number but those of higher CONV layers are smaller in size but larger in number. The size of feature maps of first CONV layer is 224x224 and their number is 64, whereas the size of feature maps in last CONV layer is 14x14 and their number is 512. This structure encodes hierarchy of features. The last three layers are FC layers, which process the features with linear operation followed by ReLU non-linearity. The output of each of the first two FC layers is a 4,096-dimensional vector and the last FC layer yeilds a vector of dimension 2,622, which is passed to a softmax layer to compute the class posterior probabilities. The input to the network is an RGB image of size 224×224. This network was trained using a huge dataset (2.6M face images of 2.6K people).

Activations of different layers of a CNN model encode very rich information and can be employed for extracting discriminative features. CONV layers encode low, medium and high level features along with their spatial information, whereas fully connected (FC) layers embed global features discarding the spatial information. Layer selection is the key success factor to extract discriminative features for a certain application. We propose to extract two types of features: (1) global, extracted from a FC layer (2) local, extracted from a CONV layer. The detail is given in the following subsections.

### 2.1.1. Local Deep CNN Features

Most of the pre-trained CNN based feature extraction methods use the activations of FC layers as representation; only few works use the activations of CONV layers [17, 18]. Unlike FC layers, CONV layers encode hierarchy of local features and retain the spatial information and can lead to more discriminative representations.

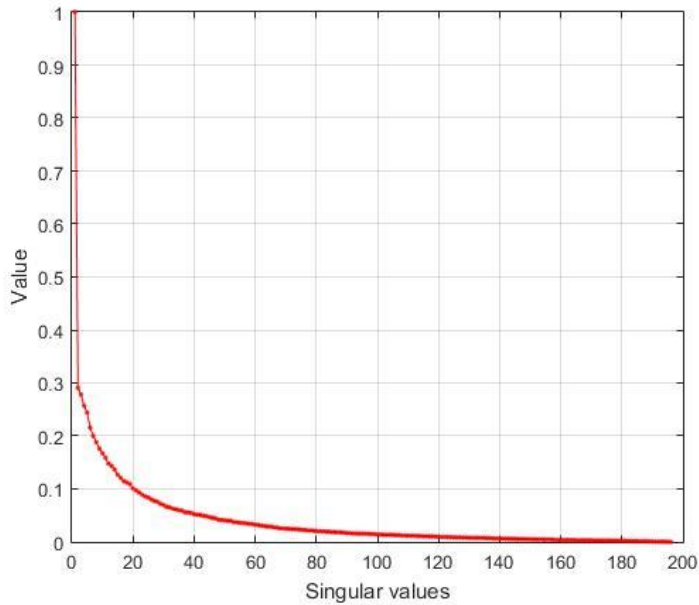

**Figure 5.** The singular value decomposition of the feature maps of an image.

For local features, we focus on CONV layers. In very deep VGG faces, low level CONV layers encode low level facial features such as micro-texture features, and high level CONV layers encode facial parts like lips, nose and different parts of periocular region. As such high CONV layers (Conv5-1, Conv5-2, Conv5-3) are the best choices for extracting local deep features. One simple way is to vectorise each feature map $p_i$ ($i = 1, 2, \ldots, n$) of a CONV layer to $x_i \in R^d$ ($d = n_1 \times n_2$, the size of $p_i$), and then concatenate them into a feature vector $x = [x_1^T, x_2^T, \cdots, x_n^T]^T \in R^D$ ($D = n_1 \times n_2 \times n$, $n$ being the number of feature maps in the CONV layer), as shown in Figure 6. This leads to the curse of dimensionality problem; the dimension of the feature space becomes excessively large, e.g. in Conv5-3, the total number of feature maps is 512 and the size of each feature map is 14×14, this results in a feature vector of dimension 100,352. We observe that there is a large amount of sparsity in feature maps. The Singular Value Decomposition (SVD) of the matrix $P = [x_1, x_2, \cdots, x_{512}] \in R^{196 \times 512}$, where the dimension of each vector is 196 (14×14), for Conv5-3 gives the singular values which are shown in Figure 5; the majority of the singular values are almost zero which is an indication of the sparsity in feature maps. It follows from this observation that the feature vector $x$ can be compressed significantly without losing noticeable amount of information. One simple and effective approach for compression is Principle Component Analysis (PCA). As in this case, the dimension of feature vectors is 100,352 but only a small number (≤2000) of eigenvalues of covariance matrix $C \in R^{D \times D}$ ($D = 100,352$ in this case) are significant, see Figure 7, so we employ the trick used in Eigenfaces approach to compute the principal eigenvalues and the corresponding principal eigenvectors to compute the PCA transformation matrix $M \in R^{D \times K}$, where $K$ is the number of features in the compressed feature vector $\hat{x} \in R^K$. In this case the computational complexity to compute depends on the number of training images, as the number of training images (e.g. 1582 images for left eyes of Samsung Daylight) is much less than the dimension (e.g. 100,352 for Conv5_3) of the local deep features, so the computation of *M* is efficient in respect of time and space complexity.

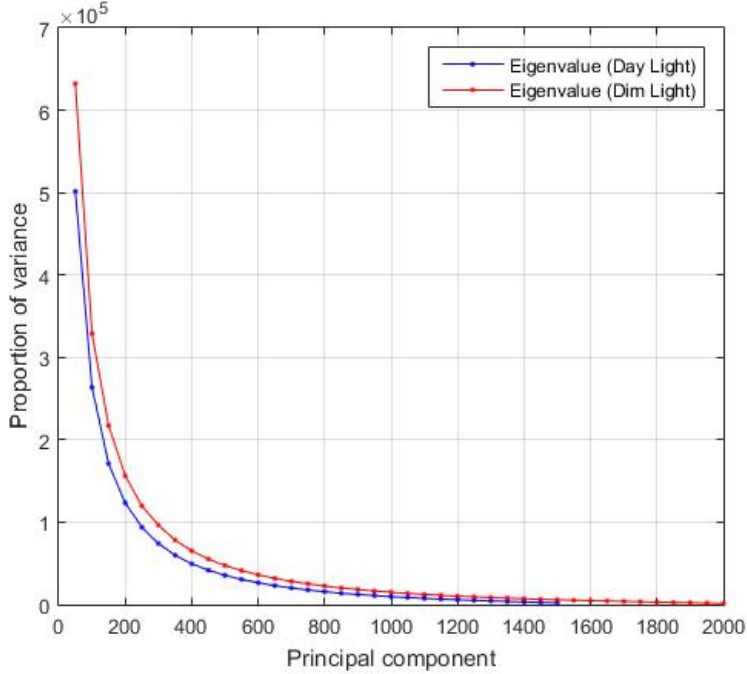

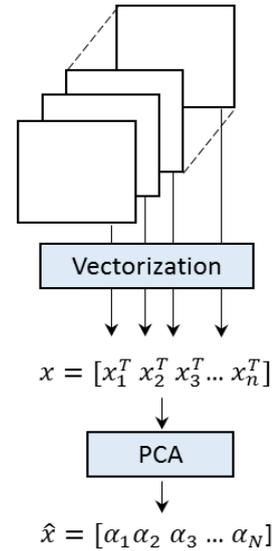

**Figure 6.** The computation of local deep CNN features.

**Figure 7.** The proportion of the variance of the principle components in Samsung (left eye) with day and dim light conditions from VISOB dataset.

### 2.1.2. Global Deep CNN Features

The activations of FC layers of VGG-Face model are taken as global features, the spatial information is not retained in FC layers. There are three FC layers in the model, we examined the effect of each layer, and the detail is provided in Section 3.1.

### 2.2. Sparsity Augmented Collaborative Representation based Classifier (SA-CRC)

Using the extracted features (global/local) from the enrolment set, we form the dictionary $\phi = [\phi_1, \phi_2, \ldots, \phi_c] \in R^{K \times N}$, where the sub-matrix $\phi_i \in R^{K \times n_i}$ is computed from feature vectors (normalized using $l_2$-norm) corresponding to $n_i$ examples of the *i*th class, *c* is the total number of classes and $N = \sum_{i=1}^{c} n_i$. In sparse representation based classification (SRC) approach, for a test example $y \in R^K$, first a collaborative representation vector $\alpha \in R^N$ is computed such that $y \approx \phi\alpha$, then class-specific residuals $r_i(y)$, ($i = 1, 2, \ldots, c$) are computed and finally *y* is assigned the label of the class for which the residual $r_i(y)$ is minimum [34]. For success of this approach, there are two different points of view: the credit goes to (i) sparseness/(ii) collaboration. The first view claims that the effectiveness of this approach is due to the sparseness of a collaborative representation vector α, whereas the other view claims that the success is due to the collaboration of the training examples from all the classes. A recent work by Naveed et al. [15] analysed that though the real gain comes from collaboration, the sparseness cannot be ignored, it further improves the classification performance. Based on this analysis they proposed SA-CRC method, which employs both sparseness and collaboration. For classification, we employed this method. In this method, first $\breve{\alpha}$ is computed by solving the problem (collaboration problem)

$$\breve{\alpha} = \min_{\alpha}\|y - \phi\alpha\|_2^2 + \lambda\|\alpha\|_2.$$

Then $\hat{\alpha}$ is calculated as the solution of the problem (sparseness problem)

$$\hat{\alpha} = \min_{\alpha}\|y - \phi\alpha\|_2^2 \ \ s.t. \|\alpha\|_0 \leq k,$$

where $k$ is the sparsity threshold. Finally these solutions are augmented

$$\alpha = \frac{\breve{\alpha} + \hat{\alpha}}{\|\breve{\alpha} + \hat{\alpha}\|_2},$$

and $y$ is assigned the label of $i$th class for which the component $q_i$ of the vector $Q = L\alpha$ is maximum, where $L$ is the label matrix. For detail, see [15].

## 3. Evaluation Protocol

We implemented the system using MATLAB R2015a on a PC with Intel (R) Core ™ i7-3610QM CPU @ 2.30 GHz and 12 GB RAM. The experiments were conducted on VISIT 1 of VISOB dataset [16], which is available in public domain. The identification performance of the system is reported in terms of accuracy, whereas the performance of verification is reported in terms of Genuine Match Rate (GMR) at False Acceptance Rate (FAR) =$10^{-2}$ and Equal Error Rate (EER). Further, the performance of the system for verification has also been presented in terms of Receiver Operating Curves (ROC).

### 3.1. Visible Light Mobile Ocular Biometric Database (VISOB)

Visible Light Mobile Ocular Biometric Database (VISOB) contains ocular images from 550 healthy adult volunteers acquired using front facing cameras of three different smartphones i.e. Samsung Note 4, iPhone 5s and Oppo N1. The Oppo and Samsung devices captured images at 1080p resolution, while the images captured by iPhone are at 720p resolution. Data was collected during two visits (Visit 1 and Visit 2), two to four weeks apart. During each visit, the volunteers were asked to take pictures using front facing cameras of the three mobile devices in two different sessions (Session 1 and Session 2) with 10 to 15 minutes apart. The volunteers used the mobile phones in normal setting, holding the devices 8 to 12 inches away from their faces. At each session, a number of images (See Table 9) were captured under three different lighting conditions: regular office light, dim ambient light, and natural daylight. This dataset was presented as a challenge dataset in ICIP2016 for periocular recognition [16]. Figure 11 shows some example images of this dataset. For experiments, only VISIT 1 dataset is available and we used this dataset for our experiments.

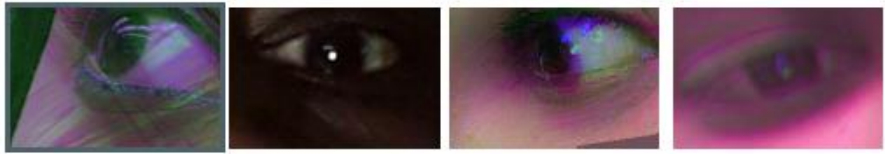

**Figure 11**. Sample images of periocular region from VISOB dataset.

**Table 9.** Detail of the enrolment and validation sets of Visit 1 and 2 of VISOB Dataset [16].

| Mobile Device | Enrollment Set (# of images) | Validation Set (# of images) |
|---|---|---|
| VISIT 1 | | |
| iPhone | 14077 | 13208 |
| Oppo | 21976 | 21349 |
| Samsung | 12197 | 12240 |
| VISIT 2 | | |
| iPhone | 12222 | 11740 |
| Oppo | 10438 | 9857 |
| Samsung | 9284 | 9548 |

## 4. Model selection

ConvSRC involves different parameters: global vs local deep features, and the number of principal components (PCs). In this section, we examine and discuss the effects of these parameters, the set of parameters that gives the highest recognition performance is chosen for further experiments. For model selection, the experiments are conducted on one set of VISOB dataset i.e. Samsung left periocular images under day light condition.

### 4.1. Global vs Local Deep Features

First, we address two main questions: which CONV/FC layer is suitable for extracting local/global deep CNN features? Which type of features (local/global) is more discriminative for periocular recognition? To answer these questions, we conducted extensive experiments for identification and verification tasks

#### 4.1.1. Global Deep CNN Features

VGG-Face contains three fully connected layers: fc6, fc7 and fc8 (see Figure 4). Fc6 and fc7 yield 4096 features whereas fc8 result in only 2622 features. The verification and identification results shown in Table 2 and Table 3, respectively, indicate that the features extracted from fc6 provide the best results. Also we notice from Table 3 that using PCA for dimensionality reduction slightly decreases the results but reduces the dimension of feature vector considerably (from 4096 to 500). We tested with fixed number (500) of principal components (PCs) and the number of PCs, which save 99% of the information.

Table 2. Verification results using features extracted from fc6, fc7 and fc8 and SA_CRC.

| Layer | Size-FV | Results | | | |
|---|---|---|---|---|---|
| | | EER | GMR at FAR=0.1 | GMR at FAR=0.01 | GMR at FAR=0.001 |
| **fc6** | **4096** | **0.88** | **99.13** | **98.32** | **94.42** |
| fc7 | 4096 | 1.21 | 98.66 | 96.77 | 85.47 |
| fc8 | 2622 | 1.89 | 97.65 | 96.1 | 77.81 |

Table 3. Identification accuracies (%) using features extracted from fc6, fc7 and fc8.

| Layer | Size-FV | Accuracy (%) | |
|---|---|---|---|
| | | KNN( Euc,1) | SA_CRC |
| fc6 | PCA-500 | 94.21 | 97.11 |
| fc6 | PCA-394 (save 99%) | 94.21 | 96.83 |
| **fc6** | **4096** | **94.21** | **97.17** |
| fc7 | 4096 | 93.54 | 95.96 |
| fc8 | 2622 | 92.62 | 93.96 |

### 4.1.2. Local Deep CNN Features (ConvSRC)

There are thirteen CONV layers in VGG-Face architecture (see Figure 4). We select the last three CONV layers (i.e. conv5_1, conv5_2 and conv5_3) for our experiments on local deep CNN features. The reason of this selection is that the first layers encode texture information whereas last layers capture higher level features which represent the rich object based structural information in a better way and result in discriminative description. Each of these CONV layer contains 512 feature maps of size 14 by 14 units. The results obtained with local features extracted using the local deep CNN feature extraction technique described in Section 2.1.1 are shown in Tables 4 and 5; these results indicate that conv5_2 outperforms the other two layers in terms of EER and GMR, there is significant difference, especially, in EER and GMR at FAR = 0.001. PCA reduces the dimension of the feature space significantly without declining the verification and identification performance as is obvious from Table 5, the feature vectors of dimensions 100352 and 500 (reduced using PCA) from Conv_5_2 give almost the same rank-1 identification result.

There is significant difference in the performance of the three CONV layers. As we move higher in the hierarchy of the CONV layers, the low level features are composed into higher level features, the results indicate that the composition of features at Conv5_2 results in the most discriminative features, which are further combined into higher level features by

Conv5_3 that create confusion. This trend further goes up to FC layers, and one can see that the performance of FC layers is even worse that Conv5_3, look at EER values with Conv5_3 in Table 5 and those in Table 2 with fac6, fc7, fc8. It indicates that Conv5_2 composes the lower level features into the representation that keep the structural information, which is important for discrimination of periocular region. From now onward, ConvSRC means the periocular recognition method that employs Conv5_2 for feature extraction.

**Table 4.** Verification results of features extracted from conv5_1, conv5_2 and conv5_3

| Layer | Size-FV | Results | | | |
|---|---|---|---|---|---|
| | | EER | GMR at FAR=0.1 | GMR at FAR=0.01 | GMR at FAR=0.001 |
| Conv5_3 | PCA-980 | 0.74 | 99.66 | 98.79 | 92.74 |
| **Conv5_2** | **PCA-1300** | **0.27** | **100** | **99.59** | **96.08** |
| Conv5_1 | PCA-876 | 0.81 | 99.33 | 98.59 | 91.12 |

**Table 5.** Identification accuracies (%) of features extracted from conv5_1, conv5_2 and conv5_3

| Layer | Size-FV | Accuracy (%) | |
|---|---|---|---|
| | | KNN( Euc,1) | SA_CRC |
| Conv5_3 | 100352 | 96.1 | 97.982 |
| **Conv5_2** | **100352** | **96.167** | **98.184** |
| **Conv5_2** | **PCA-500** | **96.01** | **97.915** |
| Conv5_1 | 100352 | 95.494 | 97.915 |

**Table 6.** Verification results of the combination of local and global features

| Global Feature | Local Feature | Size-FV | Results | | | |
|---|---|---|---|---|---|---|
| | | | EER | GMR at FAR=0.1 | GMR at FAR=0.01 | GMR at FAR=0.001 |
| FC6 | ConvSRC | PCA-500 | 0.81 | 98.66 | 97.76 | 97.23 |

### 4.2. Combining Global and Local Features

In the previous sections, we presented and discussed the individual effects of global and local deep CNN features. The discussion revealed that local features form better representation than global features. Next question is whether the fusion of local and global features can improve the recognition performance. To address this question, we tested the effect of fusing local and global features. As fc6 and Conv5_2 result in the best performances among FC and CONV layers, we fused the features from fc6 and the local features from Conv5-2. The features are fused after standardising so that each feature has zero mean and unit variance. The results are shown in Tables 6 and 7, which indicate that there is no improvement; in stead, there is some deterioration in performance, which is due to the reason that fc6 adds the features, which do not have stronger impact on discrimination and introduce redundancy, which deteriorate the performance. A comparison between global, local and their fusion for verification and identification is given in Figures 8 and 9. It validates that fusion of global and local features does not improve the performance.

Table 7. Identification accuracies (%) of the combination of local and global features

| Global Feature | Local Feature | Size-FV | Accuracy (%) | |
|---|---|---|---|---|
| | | | KNN( Euc,1) | SA_CRC |
| FC6 | ConvSRC | PCA-500 | 96.444 | 97.646 |

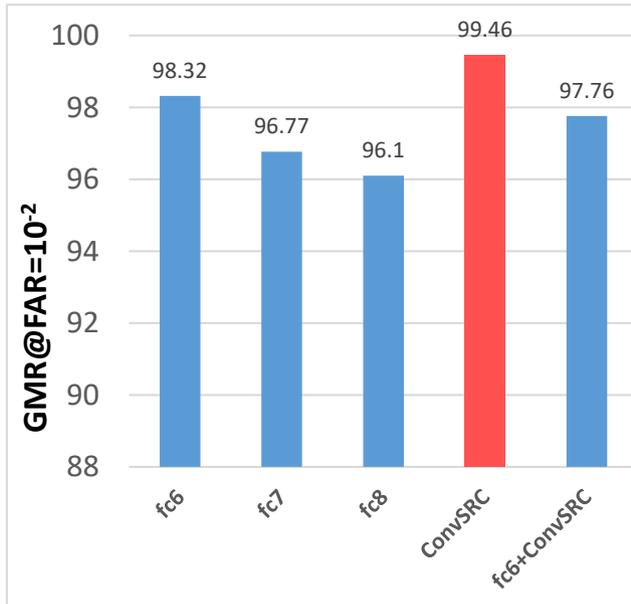

Figure 8. Comparison of the verification results using different feature extraction methods.

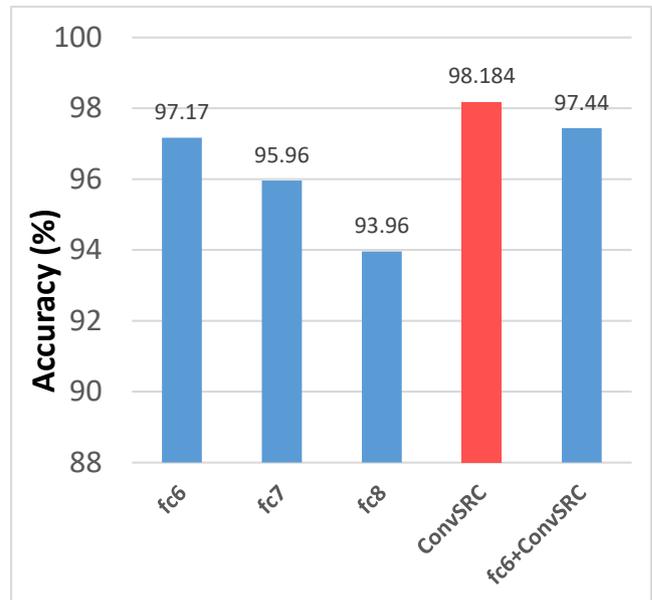

Figure 9. Comparison of the identification results using different feature extraction methods.

### 4.3. Tuning the number of principle components in PCA

The proposed approach involves dimension reduction using PCA. In PCA, the critical point is the selection of principal components (PCs) corresponding to the largest eigenvalues. Many criteria have been proposed in the literature in order to find the optimal number of dimensions in PCA. Due to the computational complexity of such techniques, we used a simple method which results in good result. We tried different numbers of components using three light conditions of the left eye of Samsung dataset. Table 8 and Figure 10 give the detail; as the number of PCs increase EER decreases and it continue to decrease until the number of PCs is 1300, where it attains minimum value for all the three cases. It is consistent with GMR, which also attains maximum value when number of PCs is 1300, see Table 8. It indicates that this number is a suitable choice for PCs, for onward experiments, we fix the number of PCs to be 1300.

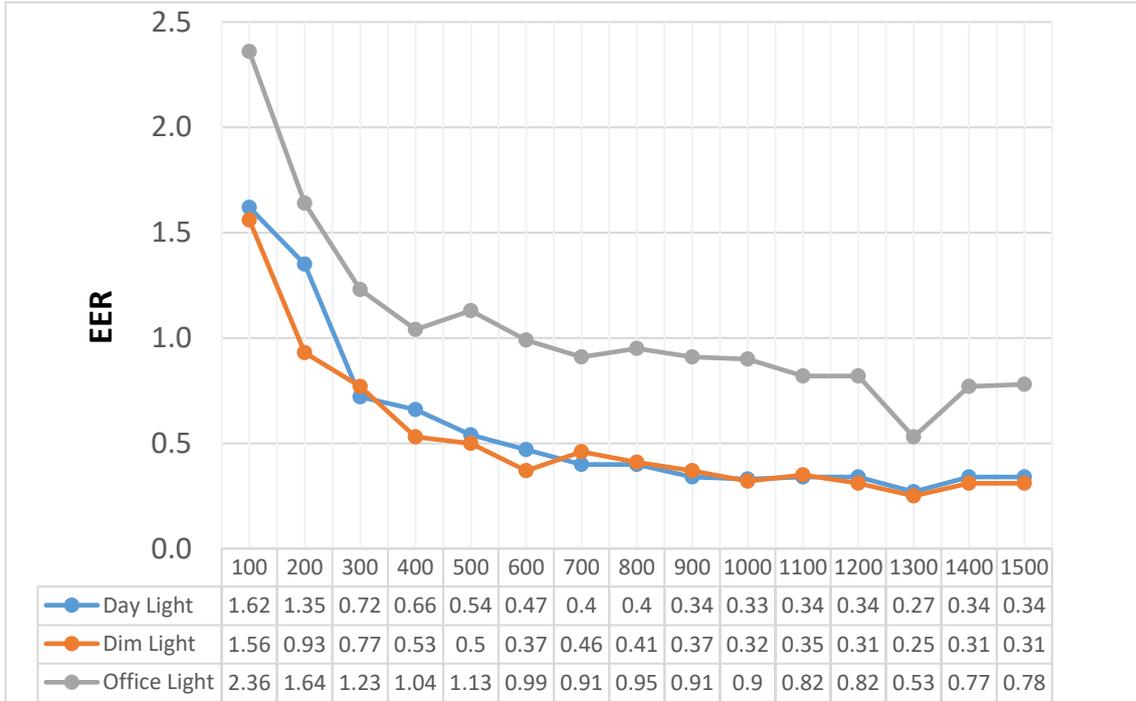

**Figure 10.** Equal Error Rate (EER) results of using different PCs on Samsung left eye dataset.

## 5. Experimental Results and Discussion

In the previous section, we presented the discussion about model selection; based on this discussion the model of the proposed method consists of local deep CNN features extracted from Conv5_2 with 1300 PCs and SA_CRC (ConvCRC). Using the model, this section reports results for verification and identification problems and discuss the performance of the proposed method in comparison with the existing methods.

### 5.1. Verification Problem

First, we report the verification results of the proposed system; Table 10 shows the performance in terms of GMR and Figures 12, 13 and 14 show the ROC curves for the left and right periocular regions captured by three types of smart phones under different lighting conditions. The results show the outstanding performance of the proposed method; it can be noted from Figures 12 to 13 that it gives better results for day-light for the three devises and these results are almost similar. The performance is slightly poor for dim-light in case of iPhone, and for office-light in case of Samsung and Oppo. Almost similar results can be noted from Table 10 in terms of GMR.

**Table 8.** The verification results using different PCs on Samsung left eye dataset.

| DB | Size-FV | GMR at FAR=0.1 | GMR at FAR=0.01 | GMR at FAR=0.001 |
|---|---|---|---|---|
| Samsung_day light | 100 | 98.25 | 97.65 | 58.24 |
| | 200 | 98.52 | 98.05 | 81.17 |
| | 300 | 99.39 | 98.39 | 89.31 |
| | 400 | 99.6 | 98.66 | 91.12 |
| | 500 | 99.6 | 99.06 | 90.32 |
| | 600 | 99.6 | 99.13 | 92.99 |
| | 700 | 99.87 | 99.19 | 92.27 |
| | 800 | 99.87 | 99.33 | 92.27 |
| | 900 | 99.87 | 99.33 | 92.47 |
| | 1000 | 99.93 | 99.19 | 92.2 |
| | 1100 | 99.87 | 99.19 | 92 |
| | 1200 | 99.93 | 99.19 | 90.99 |
| | **1300** | **100** | **99.46** | **96.08** |
| | 1400 | 99.93 | 99.26 | 93.75 |
| | 1500 | 99.93 | 99.26 | 93.48 |
| Samsung_dim light | 100 | 98.22 | 97.66 | 96.95 |
| | 200 | 99.13 | 98.63 | 97.96 |
| | 300 | 99.24 | 98.88 | 98.22 |
| | 400 | 99.64 | 99.19 | 98.42 |
| | 500 | 99.75 | 99.08 | 98.57 |
| | 600 | 99.75 | 99.29 | 98.57 |
| | 700 | 99.8 | 99.29 | 98.52 |
| | 800 | 99.8 | 99.19 | 98.63 |
| | 900 | 99.8 | 99.29 | 98.73 |
| | 1000 | 99.85 | 99.44 | 98.73 |
| | 1100 | 99.85 | 99.39 | 98.63 |
| | 1200 | 99.85 | 99.44 | 98.68 |
| | **1300** | **99.85** | **99.59** | **99.03** |
| | 1400 | 99.85 | 99.49 | 98.63 |
| | 1500 | 99.8 | 99.44 | 98.73 |
| Samsung_Office light | 100 | 96.68 | 95.68 | 91.58 |
| | 200 | 98.06 | 97.15 | 94.17 |
| | 300 | 98.58 | 97.89 | 95.04 |
| | 400 | 98.92 | 98.14 | 95.64 |
| | 500 | 98.88 | 98.1 | 95.99 |
| | 600 | 99.01 | 98.32 | 96.03 |
| | 700 | 99.09 | 98.45 | 96.29 |
| | 800 | 99.05 | 98.45 | 96.2 |
| | 900 | 99.14 | 98.58 | 96.37 |
| | 1000 | 99.14 | 98.66 | 96.25 |
| | 1100 | 99.22 | 98.66 | 96.37 |
| | 1200 | 99.22 | 98.62 | 96.33 |
| | **1300** | **99.53** | **99.01** | **97.02** |
| | 1400 | 99.71 | 98.71 | 96.37 |
| | 1500 | 99.22 | 98.79 | 96.55 |

Table 10 compares the verification performance of the state-of-the-art periocular verification methods. It can be observed that the proposed technique performs better than the state-of-the-art methods for the three mobile devices and all light conditions. It significantly outperforms MR Filters [21] and Deep Sparse Filters [22], which are the first and the second winner, respectively, of ICIP2016 periocular competition on VISOB dataset. MR Filters, Deep Sparse Filters and ConvCRC employ learning based feature extraction methods. These results show that learning based features outperform the hand-engineered features, and among learned features the features based on Deep CNN (ConvCRC) performs better because of its deeper structure, which captures the hierarchy of features in a better way and composes the discriminative description.

Moreover, we compare the performance of our method with that of MR Filters (i.e. the first winner) in detail. Figure 15 shows the detailed comparison for the three devices and the three light conditions of VISOB dataset. For fair comparison with MR filters, we reported the results in terms of GMR at FAR= $10^{-3}$ similar to that used for MR Filters. It can be observed that there is a large gap of performance that the proposed method achieved over MR Filters; the reason of better performance is that the deeper architecture of CNN captures the more discriminative description.

**Table 10.** Verification performance (GMR at FAR= $10^{-2}$) on VISOB dataset.

| Feature Type | GMR @ FMR = $10^{-2}$ | | | | | | | | | | | | | | | | | |
|---|---|---|---|---|---|---|---|---|---|---|---|---|---|---|---|---|---|---|
| | iPhone | | Oppo | | Samsung | | iPhone | | Oppo | | Samsung | | iPhone | | Oppo | | Samsung | |
| | Left | Right | Left | Right | Left | Right | Left | Right | Left | Right | Left | Right | Left | Right | Left | Right | Left | Right |
| | Capture Condition : Day Light | | | | | | Capture Condition : Dim Light | | | | | | Capture Condition : Office short Light | | | | | |
| Block BSIF | 45.77 | 42.69 | 46.22 | 49.40 | 47.63 | 48.56 | 40.05 | 35.93 | 26.62 | 51.77 | 44.44 | 48.56 | 29.47 | 30.71 | 26.62 | 23.30 | 24.45 | 30.27 |
| Block HoG | 0.11 | 0.18 | 0.35 | 0.51 | 0.09 | 0.16 | 1.19 | 1.07 | 0.49 | 0.54 | 0.22 | 0.16 | 0.36 | 0.54 | 0.49 | 0.77 | 0.10 | 0.51 |
| BSIF | 60.11 | 61.52 | 54.40 | 53.45 | 54.39 | 62.93 | 43.74 | 48.61 | 28.00 | 56.76 | 54.82 | 62.93 | 42.14 | 44.45 | 28.00 | 30.68 | 34.29 | 39.64 |
| HoG | 0.04 | 0.03 | 0.19 | 0.18 | 0.06 | 0.13 | 0.41 | 0.53 | 0.32 | 0.30 | 0.36 | 0.13 | 0.28 | 0.33 | 0.32 | 0.37 | 0.24 | 0.34 |
| LPQ | 1.65 | 1.99 | 2.75 | 2.65 | 1.41 | 9.88 | 6.70 | 7.73 | 2.88 | 5.15 | 8.70 | 9.88 | 3.53 | 2.71 | 2.88 | 3.95 | 1.96 | 1.73 |
| DSparse Filters | 93.04 | 86.26 | 96.64 | 97.56 | 90.22 | 95.03 | 89.63 | 89.47 | 87.49 | 87.08 | 91.06 | 93.18 | 88.62 | 86.62 | 87.49 | 80.09 | 79.72 | 90.92 |
| MR Filters | 92.04 | 91.34 | 92.55 | 92.70 | 93.14 | 92.29 | 92.15 | 92.92 | 93.85 | 93.98 | 93.38 | 92.64 | 88.83 | 90.08 | 93.57 | 92.49 | 89.94 | 90.63 |
| **Proposed** | **99.88** | **99.84** | **98.91** | **99.43** | **99.46** | **99.31** | **99.58** | **99.65** | **99.59** | **99.05** | **99.59** | **99.63** | **99.78** | **99.60** | **99.01** | **99.60** | **98.69** | **99.01** |

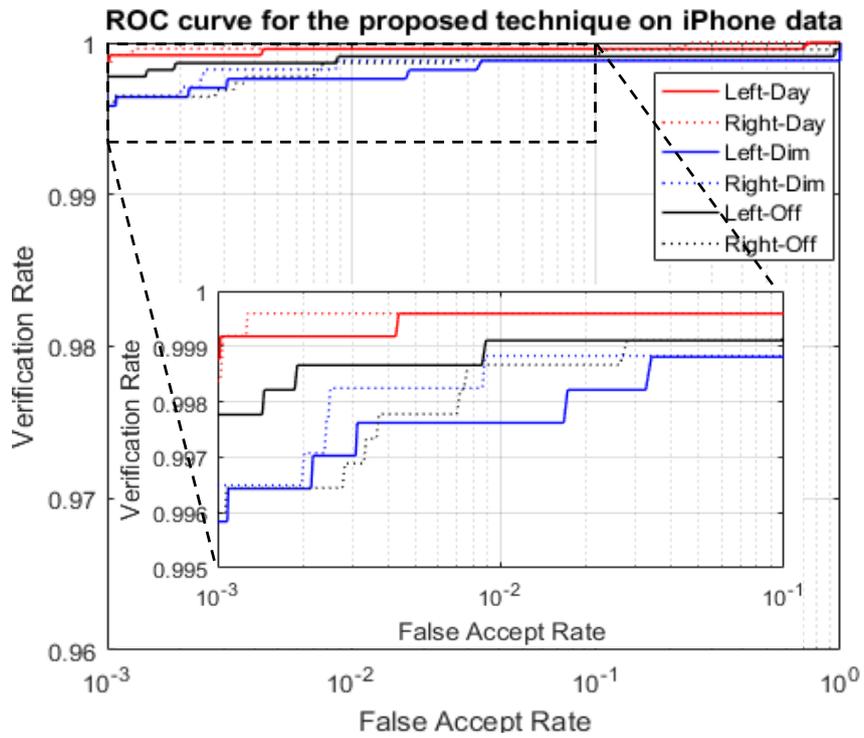

**Figure 12**. ROC curve of the proposed method on iPhone data.

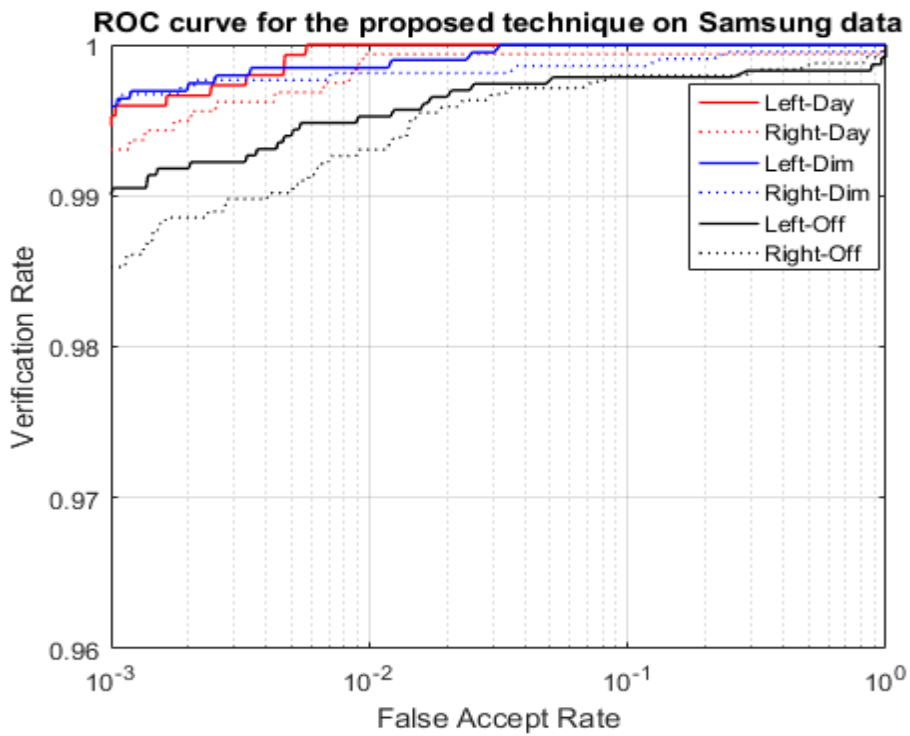

**Figure 13.** ROC curve of the proposed method on Samsung data.

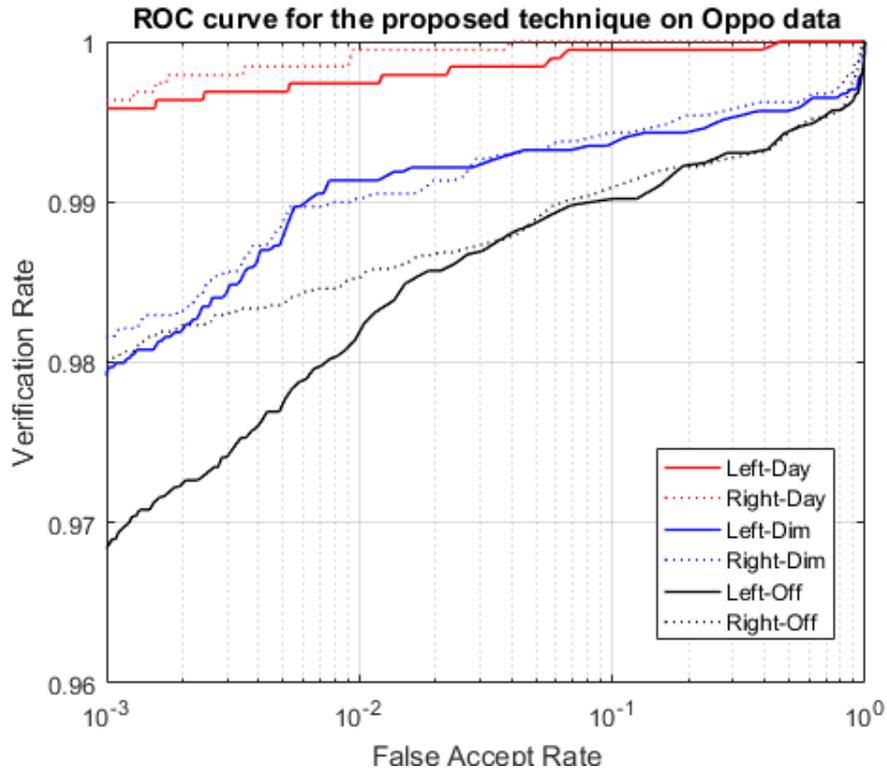

**Figure 14.** ROC curve of the proposed method on Oppo data.

## 5.2. Identification Results

VISOB dataset was prepared for verification purpose. To adapt it for identification, we only consider the images of the subjects that are available in both verification and enrolment sets. We observe from Figure 16 that the proposed method performs well for identification task. Also, we notice that the office light condition is the most challenging set in identification in case of all of the three mobile devices.

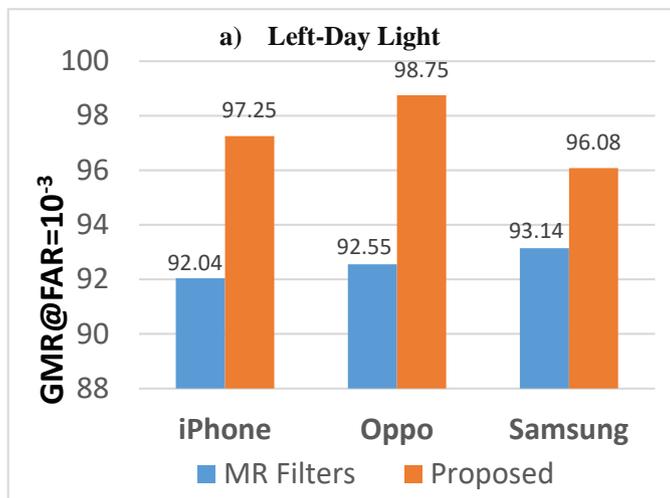
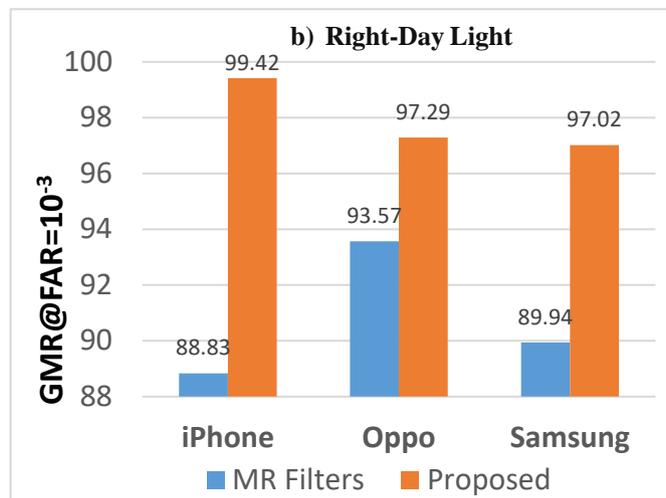
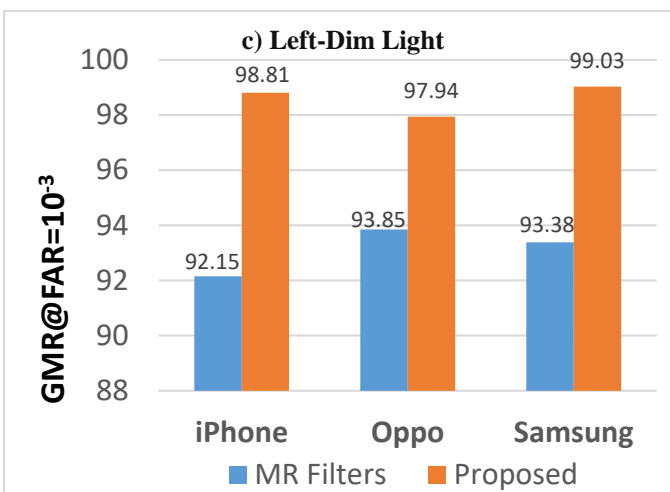
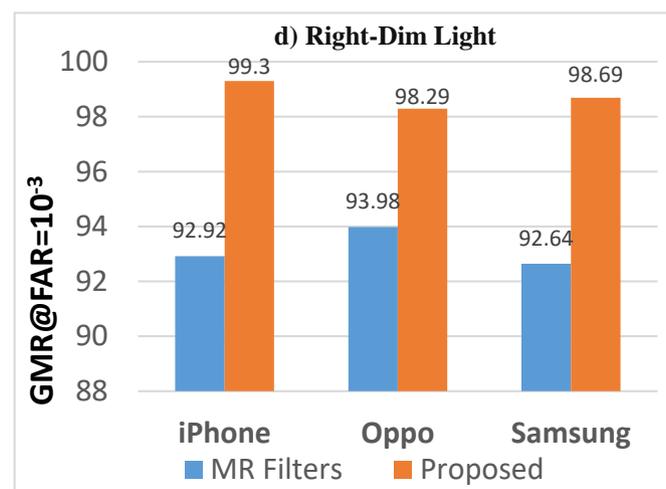
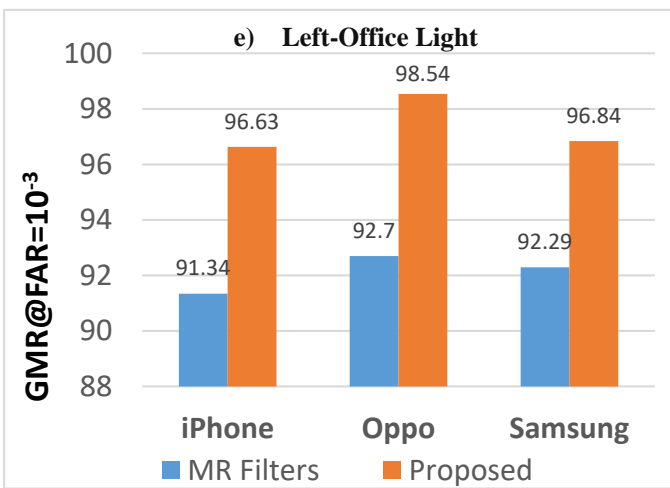
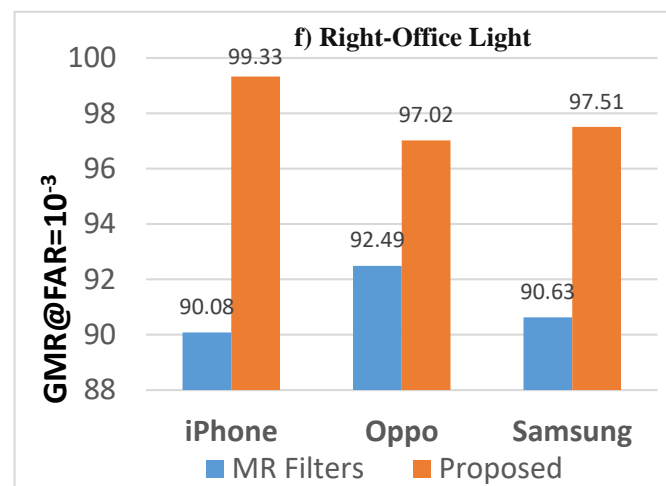

**Figure 15.** Comparison of the verification performance of the propose method with MR Filters (1st winner of ICIP2016 competition).

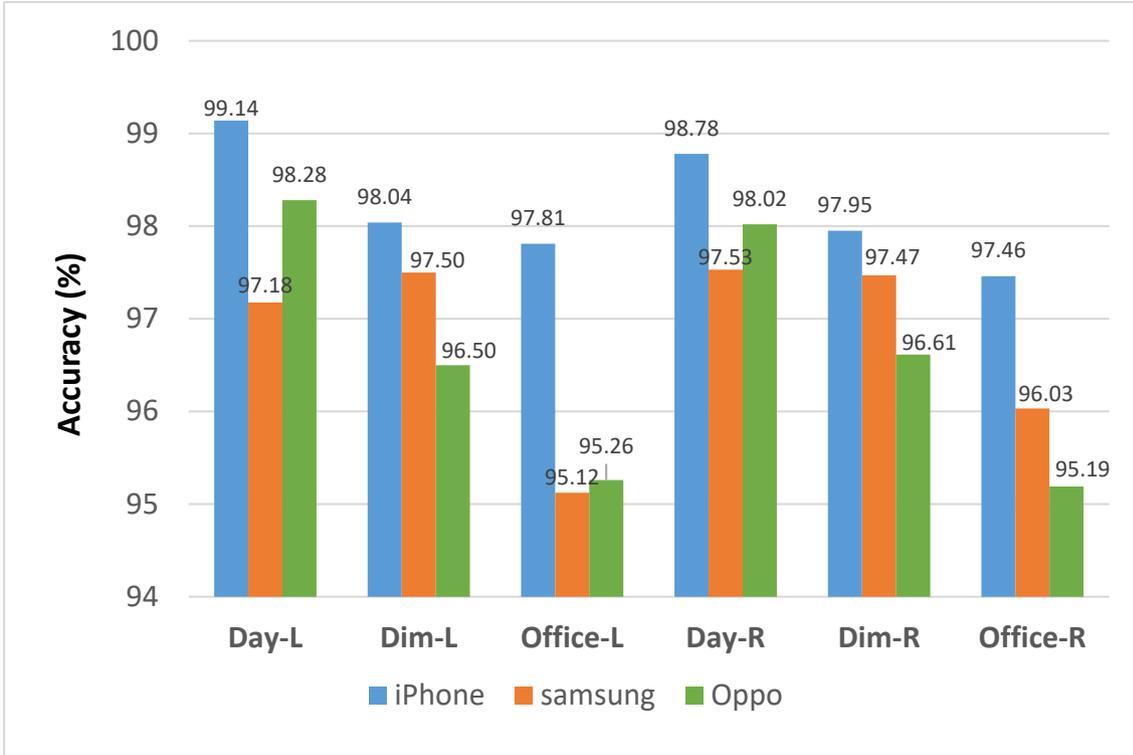

**Figure 16.** Identification performance of the proposed method on VISOB dataset.

### 5.3. Effectiveness of CNN features

The question arises whether the performance of the proposed method is due to SA-CRC classification. To this end, we passed pixel values of the original image as features to SA-CRC and compared the results with those obtained using CNN features. For this experiment we used Iphone left under day light condition dataset. The results have been shown in Table 12; there is a significant difference between the results, especially EER and GMR at FAR = $10^{-3}$. It indicates that CNN features are effective in performance improvement.

**Table 12.** Effectiveness of CNN features.

| Extraction Method | Results | | | |
|---|---|---|---|---|
| | EER | GMR at FAR=0.1 | GMR at FAR=0.01 | GMR at FAR=0.001 |
| ConvSRC | 0.08 | 99.96 | 99.88 | 97.25 |
| Original image | 0.7 | 99.47 | 98.69 | 88.57 |

### 5.4. Effectiveness of SA-CRC

The next question is about the impact of SA-CRC. For this purpose, we compare the results obtained using SA-CRC with those achieved employing KNN. The results of both classification methods on the Iphone left under day light condition dataset are given in Table 13; these results show that SA-CRC performs significantly better than KNN; there is a big gap between EER and GMR values.

Table 13. Effectiveness of SA-CRC.

| Extraction Method | Results | | | |
|---|---|---|---|---|
| | EER | GMR at FAR=0.1 | GMR at FAR=0.01 | GMR at FAR=0.001 |
| SA-CRC | 0.08 | 99.96 | 99.88 | 97.25 |
| KNN | 5.30 | 88.04 | 78.21 | 68.39 |

## 5.5. The effect of the domain of the pre-trained model

We employed pre-trained VGG-16 model, which was trained on face dataset and so it encodes the periocular regions. The question arises whether we can get the similar performance using a pre-trained VGG-16 model, which is trained on the datasets of natural images from different domain. For this we used VGG16 model that was pre-trained on ImageNet dataset [35]. The results with this model are given in Table 14 and a comparison of the results obtained with VGG16-Face and VGG16-ImageNet is depicted in Figure 17; it is surprising to note that the results are almost similar; it indicates that ConvSRC is independent of the pre-trained model domain.

Table 14. Verification performance of the proposed method (ConvSRC) using VGG16-imagnet model on VISOB dataset.

| | Phone | Light Condition | EER | GMR at FAR=0.1 | GMR at FAR=0.01 | GMR at FAR=0.001 |
|---|---|---|---|---|---|---|
| **Left** | **Samsung** | Day | 0.59 | 99.53 | 98.59 | 93.68 |
| | | Dim | 0.56 | 99.69 | 98.98 | 98.68 |
| | | Office | 0.77 | 99.22 | 98.53 | 96.55 |
| | **iPhone** | Day | 0.08 | 99.96 | 99.96 | 97.42 |
| | | Dim | 0.23 | 99.82 | 99.58 | 98.87 |
| | | Office | 0.13 | 99.96 | 99.87 | 99.42 |
| | **Oppo** | Day | 0.42 | 99.69 | 99.48 | 98.65 |
| | | Dim | 0.71 | 99.4 | 98.94 | 97.86 |
| | | Office | 0.89 | 99.16 | 98.24 | 97.02 |
| **Right** | **Samsung** | Day | 0.44 | 99.81 | 99.18 | 97.03 |
| | | Dim | 0.36 | 99.91 | 99.2 | 98.74 |

|  |  | Office | 0.66 | 99.47 | 98.65 | 97.38 |
|  | **iPhone** | Day | 0.12 | 99.92 | 99.84 | 96.87 |
|  |  | Dim | 0.18 | 99.94 | 99.65 | 99.24 |
|  |  | Office | 0.22 | 99.91 | 99.78 | 99.42 |
|  | **Oppo** | Day | 0.42 | 99.74 | 99.43 | 98.54 |
|  |  | Dim | 0.56 | 99.51 | 98.78 | 98.21 |
|  |  | Office | 1.2 | 98.71 | 98.01 | 96.96 |

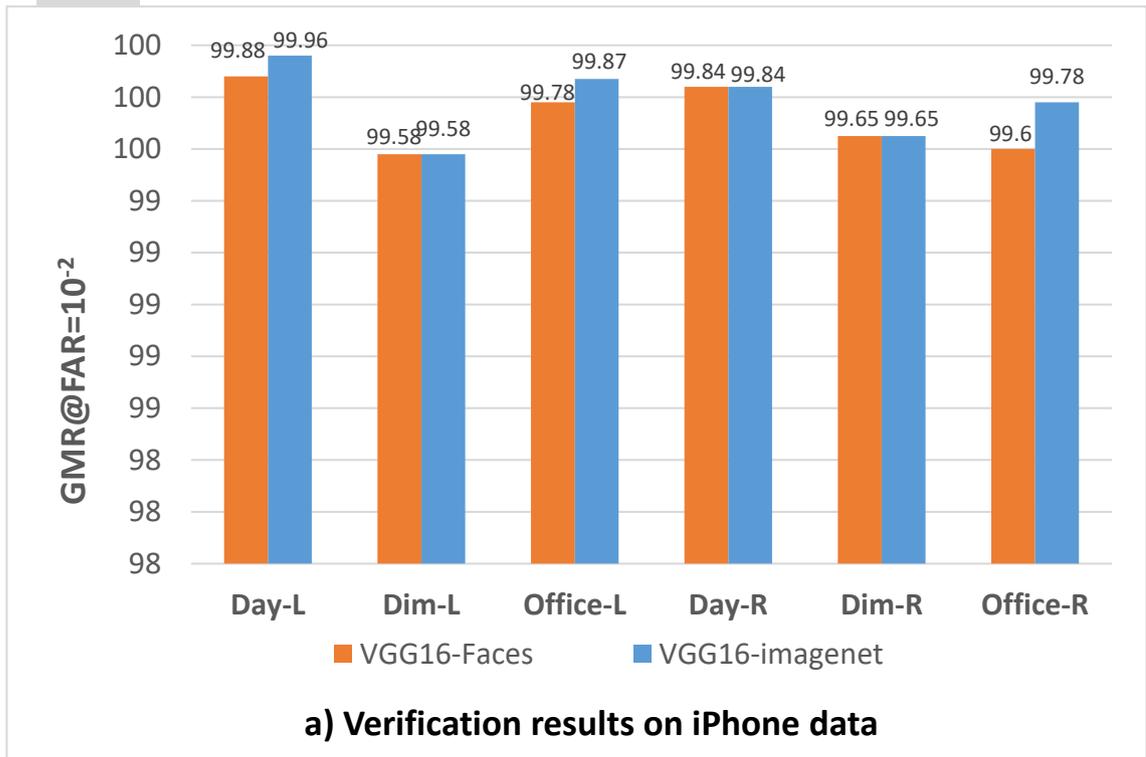

a) Verification results on iPhone data

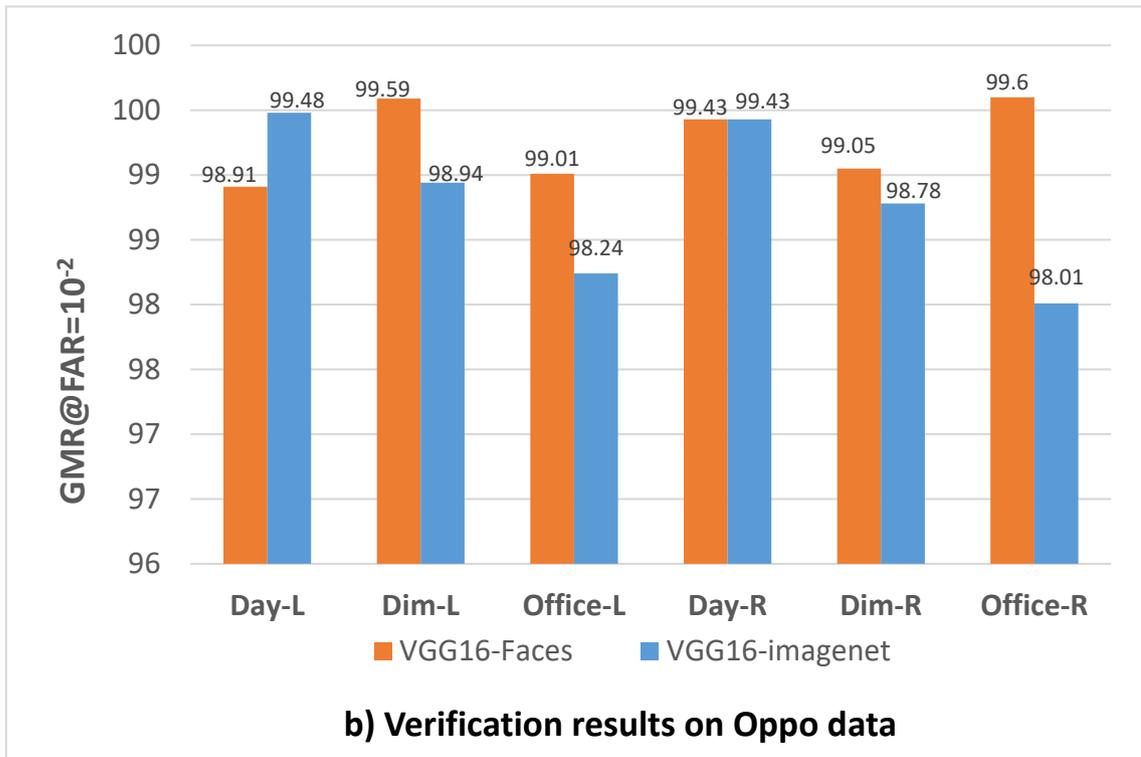

b) Verification results on Oppo data

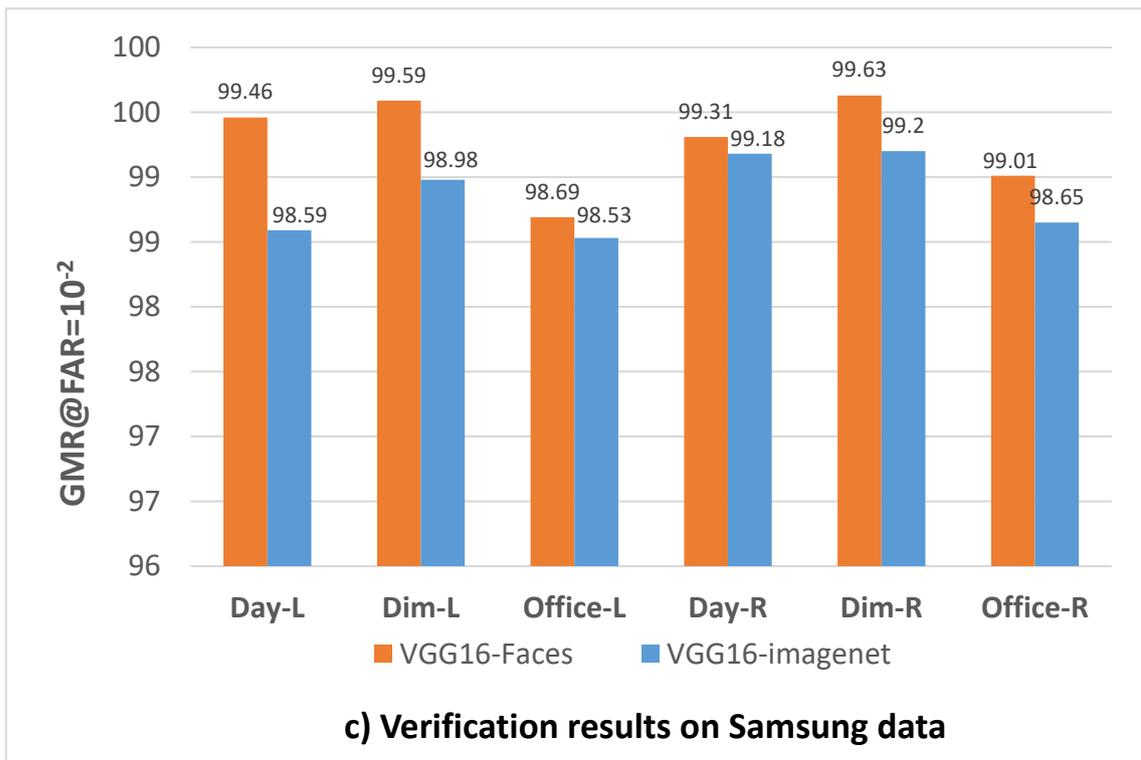

c) Verification results on Samsung data

**Figure 17.** Comparison of the verification performance of the propose method (ConvSRC) using **VGG16-Face** and **VGG16-imagnet** pre-trained models.

## 6. Conclusion

We proposed a deep learning based periocular recognition, which employs a pre-trained CNN model for discriminative feature extraction and Sparsity Augmented Collaborative Representation based Classifier (SA-CRC). For feature extraction, taking into account the wealth of information and sparsity embedded in the activations of the convolutional layers and using principle component analysis, an efficient and robust method has been proposed. We evaluated the performance of the system using convolutional layers and fully connected layers for feature extraction and found that features extracted from convolutional layers are more discriminative and robust than those obtained from FC layers; the convolutional layers at the last level result in the most discriminative representation. Through extensive experiments we have shown that pre-trained CNN features can be generalized well to periocular recognition task (verification and identification). To determine the impact of domain on pertained CNN model, we examined the performance of the system using two different models VGG-Face (pertained on face data) and VGG-Net (pertained on ImageNet data); the results indicate that the system gives eqully good performance when we use a CNN model pre-trained on any related domain i.e. ConvSRC is independent of the domain of the trained CNN model. The use of SA-CRC classifier plays a vital rule in the performance of the proposed method. We compared usefulness of SA-CRC method with stat-of-the-art KNN method; the results pointed out that SA-CRC stands out in recognition performance. The comparison with stat-of-the-art methods reveals that ConvSRC outperforms significantly even the winner of ICIP2016 completion with GMR of over 99% at FMR = $10^{-3}$.